\pdfoutput=1
\documentclass[runningheads, envcountsame, a4paper]{llncs}
\usepackage{mathtools}
\usepackage{bm}
\usepackage[export]{adjustbox}
\usepackage{subcaption}
\usepackage{threeparttable}
\usepackage{enumitem}
\usepackage{amsmath}
\usepackage{amssymb}
\setlist[itemize]{noitemsep, topsep=0pt}
\setlist[enumerate]{noitemsep, topsep=0pt}

\usepackage{enumitem}
\usepackage{mathtools}
\usepackage{amsmath}

\usepackage{subcaption}

\usepackage{algorithm}
\usepackage{algorithmic}
\usepackage{booktabs} 
\usepackage{multirow}
\usepackage{pgfplots}
\pgfplotsset{compat=1.14}

\usepackage[colorlinks]{hyperref}

\usepackage{graphicx}
\usepackage[misc]{ifsym}

\newcommand{\repeatthanks}{\textsuperscript{\thefootnote}}

\begin{document}

\title{Metapath and Entity-Aware Graph Neural Network for Recommendation}

\titlerunning{Metapath and Entity-Aware Graph Neural Network for Recommendation}
\toctitle {Metapath and Entity-Aware Graph Neural Network for Recommendation}

\author{
Muhammad Umer Anwaar (\Letter) \inst{1,3} \thanks{Equal Contribution}
\and
Zhiwei Han\inst{1,2} \repeatthanks \and
Shyam Arumugaswamy  \inst{1,3} \repeatthanks
\and
Rayyan Ahmad Khan  \inst{1,3} \repeatthanks
\and
Thomas Weber\inst{4} \and
Tianming Qiu\inst{2} \and Hao Shen\inst{2} \and Yuanting Liu\inst{2} \and
Martin Kleinsteuber\inst{1,3}}
\institute{Technische Universit{\"a}t M{\"u}nchen, Munich, Germany \and
fortiss GmbH, Munich, Germany\and
Mercateo, Munich, Germany \and
Ludwig-Maximilians-University, Munich, Germany \\
\email{
umer.anwaar@tum.de
}}

\authorrunning{Anwaar et al.}


\maketitle              

\begin{abstract}
In graph neural networks (GNNs), message passing iteratively aggregates nodes' information from their direct neighbors while neglecting the sequential nature of multi-hop node connections.
Such sequential node connections e.g., metapaths,  capture critical insights for downstream tasks.
Concretely, in recommender systems (RSs), disregarding these insights leads to inadequate distillation of collaborative signals. 
In this paper, we employ collaborative subgraphs (CSGs) and metapaths to form metapath-aware subgraphs, which explicitly capture sequential semantics in graph structures.
We propose meta\textbf{P}ath and \textbf{E}ntity-\textbf{A}ware \textbf{G}raph \textbf{N}eural \textbf{N}etwork (PEAGNN), which trains multilayer GNNs to perform metapath-aware information aggregation on such subgraphs.
This aggregated information from different metapaths is then fused using attention mechanism. 
Finally, PEAGNN gives us the representations for node and subgraph, which can be used to train MLP for predicting score for target user-item pairs.
To leverage the local structure of CSGs, we present entity-awareness that acts as a contrastive regularizer on node embedding.
Moreover, PEAGNN can be combined with prominent layers such as GAT, GCN and GraphSage.
Our empirical evaluation shows that our proposed technique outperforms competitive baselines on several datasets for recommendation tasks.
Further analysis demonstrates that PEAGNN also learns meaningful metapath combinations from a given set of metapaths.
\end{abstract}

\section{Introduction}
Integrating content information for user preference prediction remains a challenging task in the development of recommender systems. In spite of their effectiveness, most collaborative filtering (CF) methods \cite{konstan1997grouplens,he2017aneural,sarwar2001item,su2009survey} still suffer from the incapability of modeling content information such as user profiles and item features \cite{huang2002graph,lops2011content}.
Several methods have been proposed to address this problem. 
Most of them fall in these two categories: factorization and graph-based methods.
Factorization methods such as factorization machine (FM) \cite{rendle2010factorization}, neural factorization machine (NFM) \cite{he2017bneural} and Wide\&Deep models \cite{cheng2016wide} fuse numerical features of each individual training sample. 
These methods yield competitive performance on several datasets.
However, they neglect the dependencies among the content information.
Graph-based methods such as NGCF \cite{wang2019neural}, KGAT \cite{wang2019kgat}, KGCN \cite{wang2019knowledge}, Multi-GCCF\cite{sun2019multi} and LGC \cite{he2020lightgcn} represent recommender systems with graph structured data and exploit the graph structure to enhance the node-level representations for better recommendation performance \cite{berg2017graph,wang2019neural,yu2013recommendation,shi2018heterogeneous}. 

It is to be noted that learning 
such node-level representations loses the correspondences and interactions between the content information of users and items. 
This is because the node embeddings are learned independently as indicated by Zhang et al. \cite{zhang2019inductive}.
Another disadvantage of previous GNN-based methods is that the sequential nature of connectivity relations are either ignored (Knowledge Graph based methods) or mixed up (GNN-based methods) without the explicit modelling of multi-hop structure.  

A natural solution of capturing the inter- and intra-relations between content features and user-item pairs is to explore the high-order information encoded by metapaths \cite{dong2017metapath2vec,zhang2014meta}. 
A metapath denotes a set of composite relations designed for representing multi-hop structure and sequential semantics. 
To our best knowledge, only a limited number of efforts have been made to enhance GNNs with metapaths. 
A prominent metapath based method is MAGNN \cite{fu2020magnn}: it aggregates intra-metapath information for each path instance.
As a consequence, MAGNN suffers from high memory consumption problem.
MEIRec \cite{fan2019metapath} utilizes the structural information in metapaths to improve the node-level representation for intent recommendation, but the method fails to generalize when no user intent or query is available.

To overcome these limitations,
we propose Meta\textbf{P}ath- and \textbf{E}ntity-\textbf{A}ware \textbf{G}raph \textbf{N}eural \textbf{N}etwork (PEAGNN), a unified GNN framework, which aggregates information over multiple metapath-aware subgraphs and fuse the aggregated information to obtain node representation using attention mechanism. 
As a first step, we extract an $h$-hop enclosing collaborative subgraph (CSG). Each CSG is centered at a user-item pair and aimed to suppress the influence of feature nodes from other user-item interactions.
Such local subgraphs contain rich 
semantic and collaborative information of user-item interactions. 
One major difference between the CSGs in our work and the subgraphs proposed by Zhang et al. 
\cite{zhang2019inductive} is that in the subgraphs in their work neglect side information by excluding all feature entity nodes.

As a second step, the CSG is decomposed into $\gamma$ metapath-aware subgraphs based on the schema of the selected metapaths. 
After that, PEAGNN updates the node representation of the given CSG and outputs a CSG graph-level representation, which distills the collective user-item pattern and sequential semantics encoded in the CSG.
A multi-layer perceptron is then trained to predict the recommendation score of a user-item pair.
To further exploit the local structure of CSGs, we introduce entity-awareness, a contrastive regularizer which pushes the user and item nodes closer to the connected feature entity nodes, while simultaneously pushing them apart from the unconnected ones.
PEAGNN learns by jointly minimizing Bayesian Personalized Rank (BPR) loss and entity-aware loss.
Furthermore, PEAGNN can be easily combined with any graph convolution layers such as GAT, GCN and GraphSage.

In contrast to existing metapath based approaches \cite{dong2017metapath2vec,fan2019metapath,fu2020magnn,sun2012will,zhang2014meta}, PEAGNN avoids the high computational cost of explicit metapath reconstruction.
This is achieved by metapath-guided propagation. The information is propagated along the metapaths “on the fly”. This is the primary reason of computational efficiency of PEAGNN as it gets rid of applying message passing on the recommendation graph. 
Consequently, the redundant information propagated from other interactions (subgraphs) is avoided by only performing metapath-guided propagation on individual metapath-aware subgraph. We discuss this in detail in Sec. 3.

The contributions of our work are summarized 
as follows:
\begin{enumerate}[noitemsep]
    \item We decouple the sequential semantics conveyed by metapaths into different metapath-aware subgraphs and propose PEAGNN, which explicitly propagates and aggregates multi-hop semantics on metapath-aware subgraphs.
    \item We fuse the aggregated information from metapath-aware subgraphs using attention to get representations. For a given CSG, we utilize the graph-level representation and predict the recommendation score of the target user-item pairs.
    \item We introduce entity-awareness that acts as a contrastive regularizer on the node embeddings during training.
    \item The empirical analysis on three public datasets demonstrate that PEAGNN outperforms other competitive baselines and is capable of learning meaningful metapath combinations.
\end{enumerate}

\section{Related Work}
GNN is designed for learning on graph structured data \cite{scarselli2008graph,bruna2013spectral,kipf2016semi}. GNNs employ message passing algorithm to pass messages in an iterative fashion between nodes to update node representation with the underlying graph structure.
An additional pooling layer is typically used to extract graph representation for graph-level tasks, e.g., graph classification or clustering.  
Due to its superior performance on graphs, GNNs have achieved state-of-the-art performance on node classification \cite{kipf2016semi}, graph representation learning \cite{hamilton2017inductive} and RSs \cite{ying2018graph}. 
In the task of RSs, relations such as user-item interactions and user-item features can be presented as multi-typed edges in the graphs. 
Severel recent works have proposed GNNs to solve recommendation tasks \cite{wang2019neural,wang2019kgat,berg2017graph}. 
NGCF \cite{wang2019neural} embeds bipartite graphs of users and items into node representation to capture collaborative signals. 
GCMC \cite{berg2017graph} proposed a graph auto-encoder framework, which produces latent features of users and items through a form of differentiable message passing on the user-item graph. KGAT \cite{wang2019kgat} proposed a knowledge graph based attentive propagation, which enhances the node features by modeling high-order connectivity information. 
Multi-GCCF \cite{sun2019multi}  explicitly incorporates multiple  graphs in the embedding learning process and consider the intrinsic difference between user nodes and item nodes in performing graph convolution.

Prior to GNNs, several efforts have been established to 
explicitly guide the recommender learning with 
metapaths \cite{heitmann2010using,sun2012will,yu2013collaborative}. Heitmann et al. \cite{heitmann2010using} utilized linked data from heterogeneous data source to enhance collaborative filtering for the cold-start problem. 
Sun et al. \cite{sun2012will} converted recommendation tasks to relation prediction problems and tackled it with metapath-based relation reasoning. 
Yu et al. \cite{yu2013collaborative} employed matrix factorization framework over meta-path similarity matrices to perform recommendation.
Hu et al. \cite{hu2018leveraging} proposed user- and item-metapath based co-attention to fuse metapath information for recommendation while ignored the inter-metapath interactions. Zhao et al. \cite{zhao2017meta} fed the semantics encoded by meta-graph in factorization model but neglected the contribution of each individual meta-graph.
However, only a limited number of works attempted to enhance GNNs with metapaths. Two recent works are quite prominent in this regard, MAGNN \cite{fu2020magnn} and MEIRec \cite{fan2019metapath}.
MAGNN aggregates intra-metapath information for each path instance. But this causes MAGNN to get in the problem of unaffordable memory consumption. 
MEIRec devised a GNN to perform metapath-guided propagation. But MEIRec fails to generalize when no user intent is available and does not distinguish the contribution of metapaths. In contrast to these methods, our method saves computational time and memory by adopting a stepwise information propogation over meta-path aware subgraphs.
Moreover, PEAGNN employs collaborative subgraph (CSG) to separate semantics introduced by different metapaths and fuses those semantics according to the learned metapath importance, in contrast to existing approaches \cite{zhang2019inductive,fu2020magnn,fan2019metapath}.


\section{Methodology}
\subsection{Task description}
We formulate the recommendation task as:
Given a HIN that includes user-item historical interactions as well as their feature entity nodes.
The aim is to learn heterogeneous node representations 
and the graph-level representations from given collaborative subgraphs.
The graph-level representations are utilized by a prediction model to predict the interaction score between user-item pair.

\begin{figure*}
    \centering
    \includegraphics[width=\textwidth]{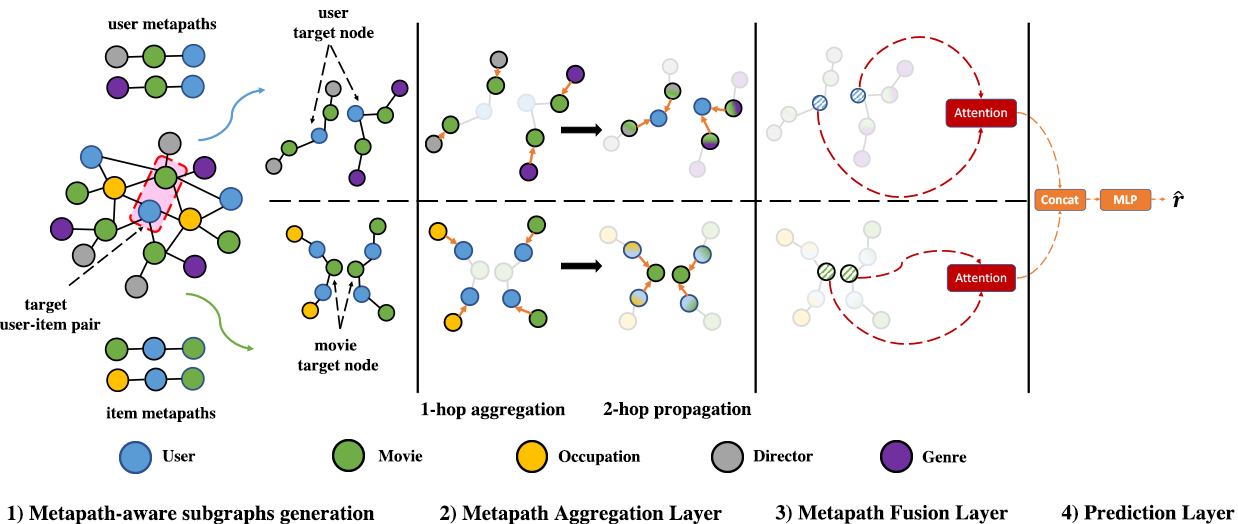}
      \vspace{-3mm}
    \setlength{\belowcaptionskip}{-8pt}   
    \caption{Illustration of the proposed PEAGNN model on the MovieLens dataset. Subfigure (1) shows the metapath-aware subgraphs generated from a CSG with the given user- and item metapaths. Subfigures (2), (3) and (4) illustrate the metapath-aware information aggregation and fusion workflow of the PEAGNN model. For simplicity, we have only adopted $2$-hop metapaths.}
    \label{fig:framework}
\end{figure*}

\subsection{Overview of PEAGNN}

PEAGNN is a unified GNN framework, which exploits and fuses rich sequential semantics in selected metapaths. 
To leverage the underlying local structure of the graph for recommendation, we introduce an entity-aware regularizer that distinguishes users and items from their unrelated features in a contrastive fashion.
Figure \ref{fig:framework} illustrates the PEAGNN framework, which consists of three components: 
\begin{enumerate}[noitemsep]
    \item A \textbf{Metapath Aggregation Layer}, which explicitly aggregates information on metapath-aware subgraphs.
    \item A \textbf{Metapath Fusion Layer}, which fuses the aggregated node representations from multiple metapath-aware subgraphs using attention mechanism.
    \item A \textbf{Prediction Layer}, which readouts the graph-level representations of CSGs and estimate the likelihood of potential user-item interactions.
\end{enumerate}

\subsection{Metapath Aggregation Layer}
Sequential semantics encoded by metapaths reveal 
different aspects towards the connected objects. 
Appropriate modelling of metapaths can improve the 
expressiveness of node representations. Our aim is 
to learn node representations that preserve the 
sequential semantics in metapaths.
PEAGNN saves memory and computation time by performing a step-wise 
information propagation over metapath-aware subgraphs.
This is contrast to Fan et al. \cite{fan2019metapath} which consider
each individual path as input.

\subsubsection{Metapath-aware Subgraph}
A metapath-aware subgraph is a directed graph induced from 
the corresponding CSG by following one 
specific metapath. 
As the goal is to learn metapath-aware 
user-item representation for recommendation,
it is intuitive to choose such metapaths 
which end with
either a user or an item node.
This ensures that the information aggregation on metapath-aware 
subgraphs always end on nodes of our primary interest. 

\subsubsection{Information Propagation on Metapath-aware Subgraphs}
\begin{figure}
 \vspace{-2mm}
    \centering
    \includegraphics[width=0.4\textwidth]{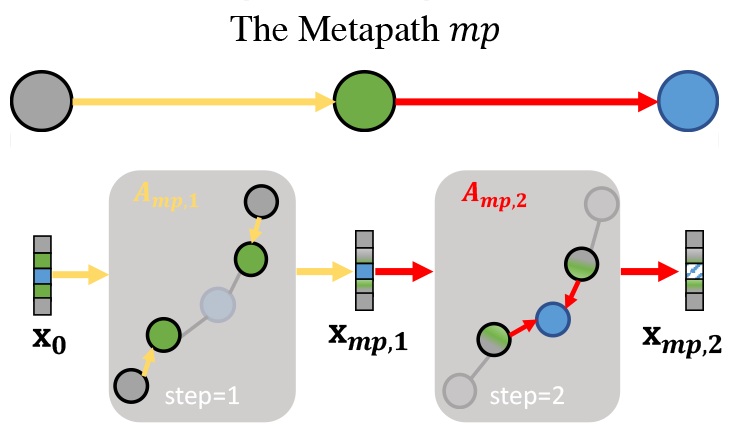}
      \vspace{-3mm}
        \setlength{\belowcaptionskip}{-8pt}  
    \caption{Information  propagation  on a metapath-aware subgraph} 
    \label{fig:propagation}
    \vspace{-1mm}
\end{figure}

PEAGNN trains a GNN model to perform step-wise information aggregation on metapath-aware subgraphs. By stacking multiple GNN layers, PEAGNN is capable of not only explicitly exploring the multi-hop connectivity in a metapath but also capturing the collaborative signal effectively. 
Fig. \ref{fig:propagation} illustrates the flow of information propagation on a given metapath-aware subgraph generated from the metapath $mp$. 
Here, $\bm{X}_{mp, k}$ is the node representations on the metapath $mp$ after $k$th propagation. 
$\bm{A}_{mp, k}$ is the adjacency matrix of the metapath $mp$ at step $k$. 
We employ orange and red color to highlight the edges being propagated at a certain aggregation step. 
Considering the high-order semantic revealed by multi-hop metapaths, we stack multiple GNN layers and recurrently aggregate the representations on the metapaths, so that the high-order semantic is injected into node representations. The metapath-aware information aggregation is shown as follows
 \vspace{-2mm}
\begin{equation}\label{eq: propagation1}
    \bm{X}_{mp, 1}=\sigma(GNN_{mp, 1}(\bm{X}_0, \bm{A}_{mp, 1})),\\ \hspace{5mm}
    \bm{X}_{mp, 2}=\sigma(GNN_{mp, 2}(\bm{X}_{mp, 1}, \bm{A}_{mp, 2})),
\end{equation}
where $\bm{X}_0$ denotes initial node embeddings. Without loss of generality, by stacking $N$ GNN layers we take into account $N$-hop neighbours information from the metapath-aware subgraph.
Thus, the node representations in the metapath-aware subgraph are given by:
 \vspace{-2mm}
\begin{equation}\label{eq: propagation2}
    \bm{X}_{mp, n}=\sigma(GNN_{mp, n}(\bm{X}_{mp, n-1}, \bm{A}_{mp, n})).
\end{equation}
$\bm{X}_{mp}$ is the output node representation of the last step on the metapath $mp$.
\subsection{Metapath Fusion Layer} 
After information aggregation within metapath-aware subgraphs,
the metapath fusion layer combines and fuses the semantic information revealed by all metapaths. 
Assume for a node $v$, a set of its node representations $\{\mathbf{x}_{mp_1}^v, \mathbf{x}_{mp_2}^v, ..., \mathbf{x}_{mp_\gamma}^v\}$ is aggregated from $\gamma$ metapaths.
Semantics disclosed by metapaths are not of equal importance to node representations and the contribution of every metapath should also be adjusted accordingly. Therefore, we leverage soft attention to learn the importance of each metapath, instead of adopting element-wise $mean$, $max$ and $add$ operators.
It is to be noted that PEAGNN applies a node-wise attentive fusion of metapath aggregated node representation.
This is contrast to previous works which employ a fixed attention factor for all nodes. Consequently, they fail to capture the node-specific metapath preference.
For a given target node $v$, we apply vector concatenation on its representations from $\gamma$ metapath-aware subgraphs, denoted as $\mathbf{H}_v=[\mathbf{x}_{mp_1}^v;\mathbf{x}_{mp_2}^v;...; \mathbf{x}_{mp_\gamma}^v]$.
The metapath fusion is performed as follows:
\begin{equation}    
\mathbf{c}_v = trace(\mathbf{W}^T \mathbf{H}_v),
\end{equation}
where $\mathbf{c}_v$ is a vector of metapath importance and $\mathbf{W}$ is a matrix with learnable parameters. We then normalize the metapath importance score using softmax function and get the attention factor for each metapath:
\begin{equation} \label{eq: fusion1}
    \text{att}_{mp_i}^v =\frac{\text{exp}(c_v^{mp_i})}{\sum_{j=1}^{\gamma}\text{exp}(c_v^{mp_j})},
\end{equation}
where $\text{att}_{mp_i}^v$ denotes the normalized attention factor of metapath $mp_i$ on the node $v$.
With the learned attention factors, we can fuse all metapath aggregated node representations to the final metapath-aware node representation, $\mathbf{e}_v$, as:
 \vspace{-2mm}
\begin{equation} \label{eq: fusion2}
    \mathbf{e}_v = \sum_{i=1}^{\gamma}\text{att}_{mp_i}^v\mathbf{x}_{mp_i}^v.
\end{equation}
\subsection{Prediction Layer}
Next, we readout the node representations of CSGs into a graph-level feature vector. 
In existing works, many pooling methods were investigated such as SumPool, MeanPooling, SortPooling \cite{zhang2018end} and DiffPooling \cite{ying2018hierarchical}. 
However, we adopt a different pooling strategy which concatenates the aggregated representations of the center user node $\mathbf{e}_u$ and item node $\mathbf{e}_i$ in the CSGs. i.e.,
\begin{equation} \label{eq: gl}
    \mathbf{e_g} = \text{concat}(\mathbf{e}_u, \mathbf{e}_i).
\end{equation}

After obtaining the graph-level representation of CSG, we utilize a 2-layer MLP to compute the matching score of a user-item pair. Lets denote a CSG with $\mathcal{G}_{u, i}$, the prediction function for the interaction score of user $u$ and item $i$ can be expressed as follows
\begin{equation} \label{eq: score}
    \Tilde{r}(\mathcal{G}_{u, i}) = \mathbf{w}_2^T\sigma(\mathbf{w}_1^T\mathbf{e}_g+\mathbf{b}_1) + \mathbf{b}_2 ,
\end{equation}

where $\mathbf{w}_1$, $\mathbf{w}_2$, $\mathbf{b}_1$ and $\mathbf{b}_2$ are the trainable parameters of the MLP which map the graph-level representation $\mathbf{e}_g$ to a scalar matching score, and $\sigma$ is the non-linear activation function (e.g. ReLU).

\subsection{Graph-level representation for recommendation}
Compared to the previous GNN-based methods such as NGCF, KGAT, KGCN, Multi-GCCF and LGC that use node-level representations for recommendation, PEAGNN predicts the matching score of a user-item pair by mapping its corresponding metapath-aware subgraph to a scalar as shown in Fig. \ref{fig:framework}. 
As shown by \cite{zhang2019inductive}, methods using node-level representation suffer from the over-smoothness problem \cite{li2018deeper}\cite{kipf2016semi}.
As their node-level representations are learned independently, 
they fail to model the correspondence of the structural proximity of a node pair. 
On the other hand, a graph-level GNN with sufficient rounds of message passing can better capture the interactions between the local structures of two nodes \cite{xu2018powerful}.

\begin{algorithm}[tb]
   \caption{The training algorithm of PEAGNN}
   \label{alg:peagat}
\begin{algorithmic}[1]
   \STATE {\bfseries Input:} A HIN $G$, 
   number of iterations $M$, $\gamma$-metapaths
   \STATE {\bfseries Output:} Learned node representations $\bm{E}$, a prediction  model $\Tilde{r}$
   \FOR{$i=1$ {\bfseries to} $M$}
        \STATE Sample a batch $\mathcal{B}$ of training user-item interactions
        \STATE Construct the CSGs of $\mathcal{B}$ as illustrated in the figure \ref{fig:framework}
        \STATE Sample the Metapath-aware Subgraphs
        from the constructed CSGs with the given $\gamma$ metapaths
        \STATE Perform step-wise information propagation over the Metapath-aware Subgraphs as per eqs. \ref{eq: propagation1} and \ref{eq: propagation2}
        \STATE For each node, obtain the metapath-aware node representation by fusing the representation from $\gamma$ metapaths with attention factors (see eqs. \ref{eq: fusion1} and \ref{eq: fusion2})
        \STATE Use the graph-level representation to calculate the interaction scores of user $u$ and item $i$ via eqs. \ref{eq: gl} and \ref{eq: score}
        \STATE Compute the training loss with eqs. \ref{eq: cf_loss}, \ref{eq: entity_loss} and \ref{eq:overall} and update model parameters
   \ENDFOR
\end{algorithmic}
\end{algorithm}

\subsection{Training Objective}
To train model parameters in an end-to-end manner, we minimize the 
pairwise Bayesian Personalized Rank (BPR) loss \cite{rendle2012bpr}, which has 
been widely used in RSs. 
The BPR loss can be expressed as follows:
\begin{equation}\label{eq: cf_loss}
    \mathcal{L}_{CF} = \sum_{(u, i_{+}, i_{-})\in \mathcal{O}}- \text{ln} \sigma \Big(\Tilde{r}(u, i_{+}) - \Tilde{r}(u, i_{-})\Big),
\end{equation}
where $\mathcal{O} = \{(u, i_+, i_-)|(u, i_+)\in R_+, (u, i_-) \in R_-\}$ is the training set, $R_+$ is the observed user-item interactions (positive samples) while $R_-$ is the unobserved user-item interactions (negative samples). The detailed training procedure is illustrated in the Algorithm \ref{alg:peagat}.

Although user and item representations can be derived by information aggregation and fusion on metapath-aware subgraphs, the local structural proximity of user(item) nodes are still missing. Towards this end, we propose \textbf{\emph{Entity-Awareness}} to regularize the local structural of user(item) nodes. The idea of \emph{entity-awareness} is to distinguish items or users with their unrelated feature entities in the embedding space. Specifically, \emph{entity-awareness} is a distance-based contrastive regularization term that pulls the related feature entity nodes closer to the corresponding user(item) nodes, while push the unrelated ones apart. The regularization term is defined as following: 
 \vspace{-1mm}
\begin{multline}\label{eq: entity_loss}
    \mathcal{L}_{Entity} = 
    \sum_{(u, i_{+}, i_{-})\in \mathcal{O}}-ln \sigma \Bigg[\Big(d(\mathbf{x}_u, \mathbf{x}_{f-, u}) - d(\mathbf{x}_u, \mathbf{x}_{f+, u})\Big) \\
    + \Big(d(\mathbf{x}_{i_+}, \mathbf{x}_{f-, {i_+}}) - d(\mathbf{x}_{i_+}, \mathbf{x}_{f+, {i_+}}) \Big)\Bigg],
\end{multline}
where $\mathbf{x}_{f+, u}, \mathbf{x}_{f-, u}$ denote the observed and unobserved feature entity embeddings of user $u$,  $\mathbf{x}_{f+, i_+}, \mathbf{x}_{f-, i_-}$ denote the observed and unobserved feature entity embeddings of positive item $i$ and $d(\cdot, \cdot)$ is a distance measure on the embedding space. 
The total loss is computed by the weighted sum of these two losses. It is given by:
\begin{equation}
    \label{eq:overall}
    \mathcal{L} = \mathcal{L}_{CF} + \lambda \mathcal{L}_{Entity}, 
\end{equation}
where $\lambda$ is the weight of the \emph{entity-awareness} term.
We use mini-batch Adam optimizer \cite{kingma2014adam}.
For a batch randomly sampled from training set $\mathcal{O}$, we establish their representation by performing information aggregation and fusion on their embeddings, and then update model parameters via back propagation.
\renewcommand{\thefootnote}{\arabic{footnote}}
\section{Experiments}
We evaluate the effectiveness our approach via experiments on public datasets.
Our experiments aim to address the following research questions:
\begin{itemize}[noitemsep]
    \item \textbf{RQ1}: How does PEAGNN perform compared to other 
    baseline methods?
    \item \textbf{RQ2}: How does the \emph{entity-awareness} affect the 
    performance of PEAGNN?
    \item \textbf{RQ3}: What is the impact of different metapaths in recommendation tasks?
\end{itemize}

\subsection{Experimental Settings}
\subsubsection{Datasets}
The datasets which we included in our experimental evaluation are widely used in related works. That is, Movielens \cite{berg2017graph,he2015trirank,he2017bneural} and Yelp \cite{wang2019kgat,yu2013recommendation}.
By changing the size of Movielens from small to large, we investigated the effect of dataset scale on the performance of the proposed method and the competitive baselines.
We have three datasets of different sizes, namely: MovieLens-small (small), Yelp (medium) and MovieLens-25M (large). 
The statistics of the three datasets are summarized in Table \ref{tab:datastats}.

\textbf{MovieLens\footnote[1]{\url{https://grouplens.org/datasets/movielens/}}} is widely used benchmark dataset for movie recommendation. We use small ($\sim$ 100k ratings) and 25M ($\sim$ 25 million ratings) versions of the dataset. 
We consider movies as items and ratings as interactions. 
For ML-small, we use 10-core setting i.e. each user and item will have at least 10 interactions. 
For ML-25M, we select items which have at least 10 interactions and users with 10 to 300 interactions (from 2018 on-wards) to ensure dataset quality. 

\textbf{Yelp\footnote[2]{\url{https://www.yelp.com/dataset}}} is used for business recommendations and has around 10 million interactions. Here we consider businesses as items; reviews and tips as interactions. The original dataset is highly sparse. So, to ensure dataset quality, we select items which have at least 50 interactions and users with 10 to 20 interactions.

Along with user-item interactions, we use their features as entities to build HIN graph. For MovieLens, we employ user feature of tag and item features like year, genre, actor, director, writer etc. For Yelp, we extract user features like counts of reviews, friends, fans, stars and item features like attributes, categories and counts of stars, reviews and check-ins. 

\begin{table}[t!]
 \vspace{-2mm}
\centering
\begin{tabular}{p{2cm}||p{1.9cm}|p{1.9cm}|p{0.8cm}}
 \hline
  & \textbf{MovieLens- small} & \textbf{MovieLens- 25M} & \textbf{Yelp}\\
  \hline
  \hline
    \#Nodes & 2933 & 33249 & 89252\\
    \#Users & 608 & 14982 & 60808\\
    \#Items & 2121 & 11560 & 28237\\
    \#Interactions & 79619 & 1270237 & 754425\\
 \hline
\end{tabular}
\caption{Statistics of datasets}
\label{tab:datastats}
\vspace{-6mm}
\end{table}
\subsubsection{Evaluation Strategy and Metrics}
Leaving one interaction out evaluation strategy is one of the most
commonly followed approach to evaluate recommender systems \cite{he2017bneural,rendle2012bpr,Bayer2016,hesigir2016}. 
Recent research shows that different data splits have a huge impact on the final performance\cite{shchur2019pitfalls}. 
To avoid the raised concerns in evaluating GNN based methods, we follow \cite{he2017aneural} and adapt leave-one-out evaluation for a more stable, reliable and fair comparison. 
For each user, we set the latest interacted item as the test set. 
The remaining items are employed for training. 
For each user-item positive interaction in the training set, we employ negative sampling strategy to get four negative items for that user. 
For Yelp and ML-25M, we sample randomly while for ML-small, we sample from the unseen items for each user.
We use two evaluation metrics: Hit Ratio (HR) and Normalized Discounted Cumulative Gain (NDCG). We consider only top-10 positions of the returned results.
HR@10 indicates whether the test item is present in top-10 recommendations. NDCG@10 also takes in to account the position at which the correct item appears in the recommendations \cite{he2015trirank}.
We compute the metrics for each test user and report the average score.

\subsubsection{Hyperparameter Settings}
We implemented the PEAGNN and baselines in Pytorch Geometric 1.5.0 \cite{Fey/Lenssen/2019}. 
To determine hyper-parameters of our methods, we follow the procedure proposed in \cite{he2017aneural}. For each user, one random interaction is sampled as the validation data for parameter tuning. We cross validated the batch size of [1024, 2048, 4096], the learning rate of [0.0001 ,0.0005, 0.001, 0.005] and weight of [0.03, 0.1, 0.3, 1] for entity-awareness.
For fair comparison, we employed the same embedding dimension for both PEAGNN and the baselines.
We set embedding dimension to 64 across all models and datasets.
The representation dimension is 16 for GNN-based models and hidden layer size is 64 for factorization and GNN-based models. 
We use 2-step metapaths and attention channel aggregation for PEAGNN. 
The number of metapaths, $\gamma$, for ML-small, ML-25M and Yelp are 9, 13 and 11 respectively. 

Further implementation details of PEAGNN as well as baseline models can be found in the code\footnote[3]{
\url{https://github.com/ecml-peagnn/PEAGNN}}.

\subsubsection{Baselines}
We compare PEAGNN with several kinds of competitive baselines.
\\
\textbf{1. NFM} \cite{he2017bneural} utilizes 
    neural networks to enhance high-order feature 
    interactions with non-linearity. As suggested by 
    He et al. \cite{he2017bneural}, we apply one 
    hidden layer neural network on input features.\\
\textbf{2. CFKG} \cite{ai2018learning} applies 
TransE \cite{bordes2013translating} to learn
heterogeneous node embedding  and converts 
recommendation to a link prediction problem.\\
\textbf{3. HeRec} \cite{dong2017metapath2vec} extends the matrix factorization model with the joint learning of a set of embedding fusion functions.\\
\textbf{4. Metapath2Vec} \cite{dong2017metapath2vec} utilizes a skip-gram model to update the node embeddings generataed by metapath-guided random walks. We then use a MLP to predict the matching score with the learned embeddings.\\
\textbf{5. NGCF} \cite{wang2019neural} integrates the user-item bipartite graph structure into the embedding process for 
    collaborative filtering.\\
\textbf{6. KGCN} \cite{wang2019knowledge} exploits multi-hop proximity information with a receptive field to learn user preference.\\
\textbf{7. KGAT} \cite{wang2019kgat} incorporates high-order information by performing attentive embedding propagation with the learned entity attention on knowledge graph.\\
\textbf{8. Multi-GCCF} \cite{sun2019multi} explicitly incorporates multiple
graphs in the embedding learning process. Multi-GCCF not
only models the high-order information but also integrates the
proximal information of item-item and user-user paris.\\
\textbf{9. LGC} \cite{he2020lightgcn} simplifies the design of GCN by maintaining only the neighborhood aggregation for
collaborative filtering.

\subsection{Overall Performance Comparison (\textbf{RQ1})}
Table \ref{tab:overallperf} summarizes the performance comparison of PEAGNN variants and the competitive baselines.
In the following, we discuss these results to gain some important insights into the problem.

First, we note that our proposed method PEAGNN outperforms other methods by a significant margin on all three datasets.
In particular, the performance gains achieved by PEAGNN are 7.87\%, 2.39\%, and 8.23\% w.r.t. NDCG@10 on ML-small, ML-25m and Yelp datasets, respectively.
The primary reason for outstanding performance of PEAGNN is that
none of the existing methods do explicit modelling consequently all node messages get mixed up during message passing. 
This verifies our claim that explicit modelling and fusing sequential semantics in metapath help in better learning of user-item interactions.
Moreover, the superior performance of PEAGNN also reveals the effectiveness of  modelling the local graph structure, while other GNN-based methods simply pay no attention to their structure proximity.

Second, we observe that the path-based methods that are not based on GNN significantly underperform all the GNN-based models on all three datasets.
Both Metapath2Vec+MLP and HeRec have incorporated the static node embedding using Metapath2vec \cite{dong2017metapath2vec}. 
The poor performance indicates that learning unsupervised static node embeddings have limited the power of the model to capture the complex collaborative signals and intricate content relations.

Third, we observe that CFKG and NFM prove to be strong baselines for GNN based methods especially on ML-25m and Yelp datasets. 
For instance, CFKG achieves $0.8729$ HR@10 on Yelp and is fourth best performing model.
But the performance gap with second and third best models is negligible i.e. 0.4\% and 0.2\% respectively. 
CFKG is significantly outperformed only by our method (PEAGAT$^*$) by 4.37\%.
This demonstrates that our method is able to better leverage the  information in the graph-structured data by explicitly modelling the the sequential semantics via metapath aware subgraphs.


\begin{table*}[t!]
 \vspace{-3mm}
\centering
\begin{threeparttable}
\begin{tabular}{|c|c c|c c|c c|}
 \hline
  \multirow{2}{*}{\textbf{model}} &  \multicolumn{2}{c|}{\textbf{MovieLens-small}} & \multicolumn{2}{c|}{\textbf{MovieLens-25M}} & \multicolumn{2}{c|}{\textbf{Yelp}}\\\cline{2-7}
   &\textbf{HR@10} & \textbf{NDCG@10} & \textbf{HR@10} & \textbf{NDCG@10} & \textbf{HR@10} & \textbf{NDCG@10}\\
  \hline
    NFM & 0.477 & 0.2668 & 0.8132 & 0.5347 & 0.8595 & 0.6062\\
  \hline
    CFKG & 0.4378 & 0.2381 & 0.8152 & 0.5196 & 0.8729 & 0.5826\\
  \hline
    HeRec & 0.2668 & 0.1449 & 0.607 & 0.3291 & 0.5533 & 0.3302\\
    Metapath2Vec & 0.3063 & 0.1614 & 0.7956 & 0.5051 & 0.6307 & 0.402\\
  \hline
    NGCF & 0.5016 & 0.2755 & 0.7807 & 0.4866 & 0.8068 & 0.481\\
    KGCN & 0.5132 & 0.2788 & 0.7771 & 0.4699 & 0.8125 & 0.4668\\
    KGAT & 0.5214 & 0.2846 & 0.8147 & 0.5236 & 0.8762 & 0.6136\\
    MultiGCCF&  0.5230 & 0.2836 & 0.8014 & 0.5153 & 0.8639 & 0.6120\\
    LGC  & 0.5003 & 0.2815  & 0.8081 & 0.5237 & 0.8744 & 0.6122\\
  \hline
    \hline
    PEAGCN & 0.5382 & 0.2951 & 0.8185 & 0.5344 & 0.9041 & 0.6379\\
    PEAGCN\tnote{*} & 0.5576 & 0.3036 & 0.8187 & 0.5361 & 0.9125 & 0.6443\\
    (\% improv. w.r.t. &&&&&&\\ best competitor) & 6.62\% & 6.68\% & 0.43\% & 0.26\% & 4.14\% & 5.00\%\\
  \hline    
    PEAGAT & 0.5375 & 0.2983 & 0.8249 & 0.5414 & 0.9057 & 0.6382\\
    PEAGAT\tnote{*} & 0.5477 & 0.3045 & \textbf{0.8284} & \textbf{0.5475} & \textbf{0.9128} & \textbf{0.6641}\\
    (\% improv. w.r.t. &&&&&&\\ best competitor) & 4.72\% & 6.99\% & \textbf{1.62\%} & \textbf{2.39\%} & \textbf{4.18\%} & \textbf{8.23\%}\\
  \hline       
    PEASage & 0.5444 &  0.3003 & 0.8176 & 0.5383 & 0.8772 & 0.6247\\
    PEASage\tnote{*} & \textbf{0.5609} & \textbf{0.307} & 0.8273 & 0.5462 & 0.8837 & 0.6308\\
    (\% improv. w.r.t. &&&&&&\\ best competitor) & \textbf{7.25\%} & \textbf{7.87\%} & 1.48\% & 2.15\% & 0.86\% & 2.80\%\\
  \hline
 \hline
\end{tabular}
\caption{Overall Performance Comparison. The scores are average of five runs. Bold indicates best results for the dataset. * denotes entity-awareness.}
\label{tab:overallperf}
\end{threeparttable}
\vspace{-5mm}
\end{table*}

\subsection{Effect of Entity-awareness (RQ2)}
The goal of introducing entity-awareness is to take advantage of the first-order structure of CSG, which is not well exploited by pure message passing in graph-based RSs \cite{gilmer2017neural}. 
We study the effect of entity-awareness by comparing the performances of our models with and without entity-awareness. 
The effect of entity-awareness for different base models are summarized in Table \ref{tab:overallperf}.
Generally, entity-awareness delivers consistently better performance than PEAGNN without entity-awareness. 
In particular, a more significant performance gain has been observed in the smaller dataset ML-small with a minimum improvement of 1.9\% on HR@10.
On the other hand, models with entity-awareness slightly outperform base models on the larger and denser datasets. 
It indicates that leveraging local structure on sparse datasets proves beneficial. 
Nonetheless, NDCG@10 benefits more from the entity-awareness in  comparison with HR@10 on both MovieLens and Yelp datasets. For instance, on Yelp, PEAGAT shows 4.06\% performance gain in terms of NDCG@10. 
These results signify the importance of explicit modelling of first-order relations in RSs.

\begin{table*}[t!]
 \vspace{-3mm}
\centering
\begin{tabular}{|c|c|c|c|c|c|c|c|c|c|}
 \hline
   \textbf{\#} & \textbf{U-M-U} & \textbf{M-U-M} & \textbf{Y-M-U} & \textbf{A-M-U} & \textbf{W-M-U} & \textbf{D-M-U} & \textbf{G-M-U} & \textbf{T-M-U} & \textbf{T-U-M}\\
  \hline
    1 & \textbf{-27.45} & -4.41 & \textbf{-24.3} & -2.21 & -1.9 & -0.96 & -0.96 & -2.53 & -6 \\
  \hline
     2 & \textbf{-30.31} & \textbf{-46.98} & \textbf{-11.51} & -6.37 & -7.28 & -8.18 & -5.16 & -9.71 & -5.45 \\
  \hline
     3 & -1.88 & -5.3 & \textbf{-40.62} & -4.37 & -3.12 & -0.3 & -0.63 & -4.69 & \textbf{-15.32} \\
  \hline
    4 & -6.38 & \textbf{-48.77} & \textbf{-23.63} & +1.53 & -1.85 & -0.62 & +0.3 & -2.06 & -7.67 \\
  \hline
    5 & -8.97 & \textbf{-42.24} & \textbf{-49.38} & -4.34 & -3.1 & -1.55 & -5.29 & -1.55 & -4.04 \\
 \hline
\end{tabular}
\caption{Percentage drop in the performance of PEASage on ML-small w.r.t. HR@10 when one metapth is removed during training.
Bold indicates greater than 10\% drop in performance.
(Abbreviation for nodes: U-User, M-Movie, Y-Year, A-Actor, W-Writer, D-Director, G-Genre and T-Tag)}
\label{tab:metapatheffect}
\vspace{-4mm}
\end{table*}
\subsection{Effect of Metapaths (RQ3)}
We have conducted various ablation studies, 
In order to gain insight into the effect of different metapaths 
on the performance of PEAGNN, we have conducted various experiments.
In the interest of space, we only include the results of our best model PEASage on ML-small dataset. 
We have total 9 metapaths for ML-small dataset. These metapaths are shown as columns in the Table \ref{tab:metapatheffect}.
We drop only 1 specific metapath at a time and then compare the model performance drop with the original one. 
The table summarizes the percentage performance drop as compared to the original model.
We ran the experiments 5 times with different random seeds for fair evaluation. 

First we observe that, there are some metapaths droping which results in significant decrease in the model performance.
That is, PEASage learnt three different key metapaths combinations. 
Namely: U-M-U, M-U-M and Y-M-U.
Second, we observe that each metapath is contributing something in the performance of PEASage although the effect of six metapaths is not that significant.

Third, we note that the published year of movies has the most significant impact on users’ choice. That is, the metapath Y-M-U comes out as key metapath in all 5 runs. 
Those metapaths which capture collaborative effects such as U-M-U and M-U-M take critical role in the high performance of PEASage model.
Another interesting phenomenon, which warrants further investigation, is that the tag given by users might have a complementary relation of user-item interactions as shown in the third run in the table.

These results also indicate another strength of PEAGNN, i.e., even without prior knowledge and careful selection, the effectiveness of each metapath and different metapath combinations can be verified in a convenient way by comparing attention factors or ``disabling'' specific metapath. Thus, an incremental training and metapath selection is achievable. Therefore, more insightful research for HCI community can be expected.

\section{Conclusion}
In this work, we study the necessity of incorporating sequential semantics in metapath for recommendation. Instead of mixing up multi-hop messages in graphs, we devised a unified GNN framework called PEAGNN, which explicitly performs independent information aggregation on generated metapath-aware subgraphs. 
Specifically, a metapath fusion layer is trained to learn the metapath importance and adaptively fuse the aggregated metapath semantics in an end-to-end fashion. 
We also introduce a contrastive connectivity regularizer called entity-awareness which exploits the first-order local structure of the graph.
For first-order local structure exploitation, the entity-awareness, a contrastive connectivity regularizer, is employed on user nodes and item nodes representation.
The experiments on three public datasets have shown the effectiveness of our approach in comparison to competitive baselines.


\bibliography{references}
\bibliographystyle{splncs04}

\end{document}